\pdfoutput=1

\documentclass[11pt]{article}

\usepackage[final]{acl}

\usepackage{times}
\usepackage{latexsym}

\usepackage[T1]{fontenc}

\usepackage[utf8]{inputenc}

\usepackage{microtype}

\usepackage{inconsolata}

\usepackage{graphicx}
\usepackage{tcolorbox}
\tcbuselibrary{breakable}
\usepackage{hyperref}
\usepackage{url}
\usepackage{booktabs} 
\usepackage{subcaption}
\usepackage{tablefootnote}
\usepackage{multirow}
\usepackage{array} 
\usepackage{comment}
\usepackage{adjustbox}
\usepackage{float}
\usepackage{amsmath}
\usepackage{cleveref}
\usepackage{pgfplots}
\usepackage{pgfplotstable}
\usepackage{subcaption}

\title{
FinMTEB: Finance Massive Text Embedding Benchmark}

\author{Yixuan Tang , Yi Yang \\
The Hong Kong University of Science and Technology\\
\texttt{ytangch@connect.ust.hk, imyiyang@ust.hk}
}

\begin{document}
\maketitle
\begin{abstract}
The efficacy of text embedding models in representing and retrieving information is crucial for many NLP applications, with performance significantly advanced by Large Language Models (LLMs). Despite this progress, existing benchmarks predominantly use general-purpose datasets, inadequately addressing the nuanced requirements of specialized domains like finance. To bridge this gap, we introduce the \textbf{Fin}ance \textbf{M}assive \textbf{T}ext \textbf{E}mbedding \textbf{B}enchmark (FinMTEB), a comprehensive evaluation suite specifically designed for the financial domain. FinMTEB encompasses 64 datasets across 7 task types, including classification, clustering, retrieval, pair classification, reranking, summarization, and semantic textual similarity (STS) in English and Chinese. Alongside this benchmark, we introduce Fin-E5, a state-of-the-art finance-adapted embedding model, ranking first on FinMTEB. Fin-E5 is developed by fine-tuning e5-Mistral-7B-Instruct on a novel persona-based synthetic dataset tailored for diverse financial embedding tasks. Evaluating 15 prominent embedding models on FinMTEB, we derive three key findings: (1) domain-specific models, including our Fin-E5, significantly outperform general-purpose models; (2) performance on general benchmarks is a poor predictor of success on financial tasks; and (3) surprisingly, traditional Bag-of-Words (BoW) models surpass dense embedding models on financial STS tasks. This work provides a robust benchmark for financial NLP and offers actionable insights for developing future domain-adapted embedding solutions. Both FinMTEB and Fin-E5 will be open-sourced for the research community. \footnote{Github: https://github.com/yixuantt/FinMTEB}

\end{abstract}

\section{Introduction}

\begin{figure*}
    \centering
    \includegraphics[width=.99\linewidth]{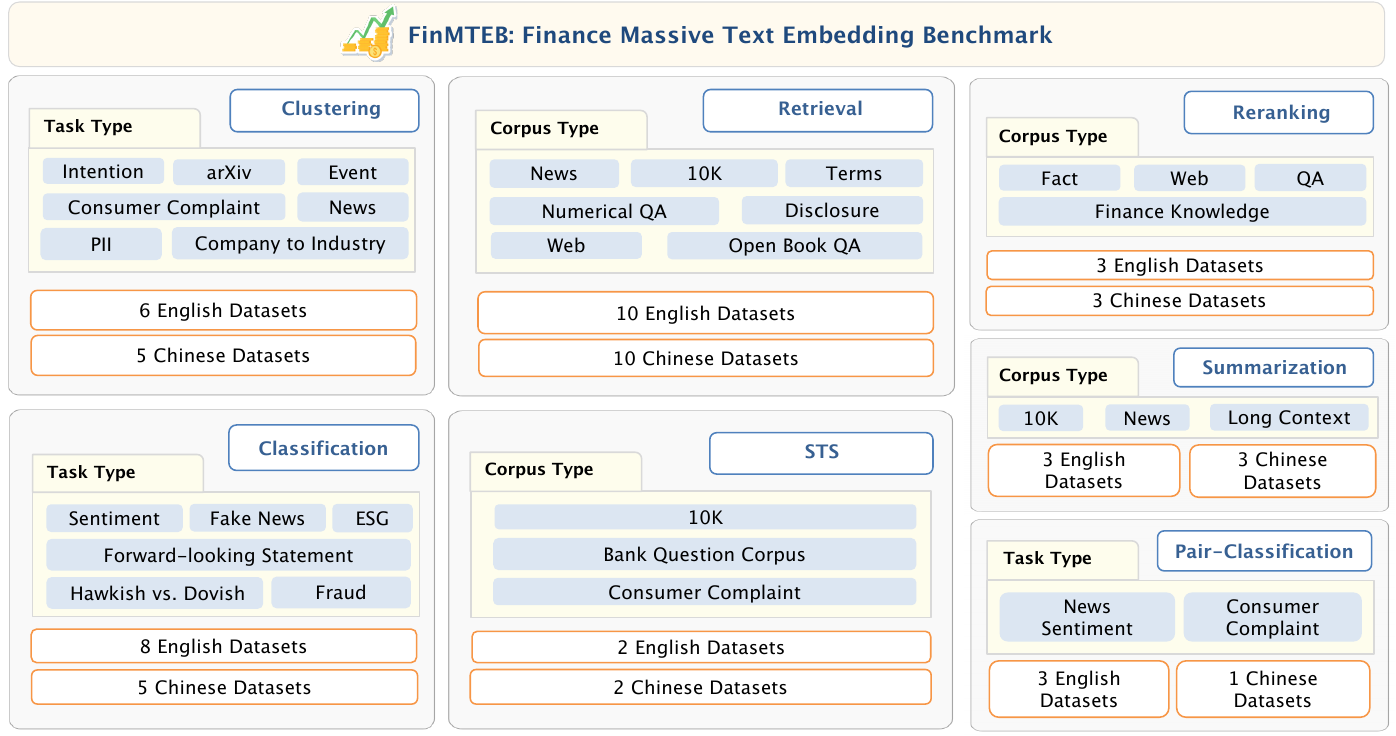}
    \caption{An overview of tasks and datasets used in FinMTEB. All the dataset descriptions and examples are provided in the Appendix \ref{append: datasets}. }
    \label{fig: overview}
\end{figure*}

Embedding models, transforming text into dense vector representations, are foundational to many natural language processing (NLP) tasks \citep{word2vector, pennington-etal-2014-glove, peters-etal-2018-deep}. Their quality significantly impacts downstream applications like information retrieval and semantic understanding. While recent Large Language Model (LLM)-based embeddings \citep{e5,gte,SFR-embedding-2} demonstrate remarkable performance on general benchmarks, their efficacy in specialized domains, particularly finance, remains under-explored. Financial text analysis presents unique challenges, including domain-specific terminology, temporal sensitivity, and complex numerical relationships \citep{alphafin,anderson-etal-2024-finance_text_embedding}, raising critical questions: \textit{How effectively do modern embedding models capture domain-specific financial information? Can domain adaptation enhance LLM-based embeddings for financial applications?}

These questions are motivated by three key insights. First, financial semantics often diverge from general language usage. For instance, "liability" inherently carries negative sentiment in financial contexts due to its association with obligations and risks, contrasting with its neutral denotation of legal responsibility in general usage. Such semantic divergence is critical for applications like Retrieval Augmented Generation (RAG) systems, where accurate document retrieval is important for effective knowledge augmentation. While recent work adapts RAG for finance \citep{alphafin,malandri-etal-2025-fin}, the fundamental role of embedding quality in retrieval efficacy is often overlooked.

Second, empirical evidence highlights the necessity of domain adaptation for optimal performance in specialized fields \citep{domainiskey,dont_stop}, even with advanced LLMs. This has led to models like BiMedLM \citep{biomedlm} for biomedical texts and BloombergGPT \citep{bloomberggpt} for finance. This specialization extends to embedding models, with examples like BioWordVec \citep{biowordvec} and FinBERT \citep{finbert}. Notably, the financial industry itself contributes to these advancements; for instance, BAM, a RoBERTa-based model from Balyasny Asset Management \citep{anderson-etal-2024-finance_text_embedding}, has demonstrated improvements. Compared to the general domain, a significant gap exists: despite commercial solutions like voyage-finance-2 \citep{voyage}, there is a lack of open-source, LLM-based financial embedding models accessible to the research community.

Third, financial NLP lacks comprehensive evaluation frameworks specifically for embedding models. Current benchmarks like FinanceBench \citep{financebench} and FinQA \citep{finqa} primarily assess text generation, while embedding-specific evaluations \citep{FiQA,finsts} are often narrow in scope, targeting single task types or limited text types. This gap is exacerbated by unique characteristics of financial texts, such as the prevalence of boilerplate language (e.g., "The company's performance is subject to various risks..."). Such standardized disclaimers, frequent but low in informational content, complicate models' ability to distinguish meaningful business insights from routine compliance text. Thus, a critical need exists for comprehensive financial embedding benchmarks.


To bridge this gap, we introduce the \textbf{Fin}ance \textbf{M}assive \textbf{T}ext \textbf{E}mbedding \textbf{B}enchmark (FinMTEB). This comprehensive benchmark comprises 64 domain-specific datasets spanning English and Chinese and covering seven critical financial embedding tasks: classification, clustering, retrieval, pair classification, reranking, summarization, and semantic textual similarity (STS). Concurrently, we develop and release Fin-E5, a finance-adapted embedding model that achieves state-of-the-art performance on FinMTEB. Fin-E5 is built by fine-tuning e5-Mistral-7B-Instruct \citep{e5} on a persona-based synthetic dataset designed to generate diverse training data relevant to various financial embedding tasks. Our extensive experiments, evaluating 15 prominent embedding models on FinMTEB, yield three crucial insights: (1) LLM-based embeddings, particularly when domain-adapted like Fin-E5, generally outperform traditional methods and their general-purpose LLM counterparts, providing significant performance gains. (2) Performance on general benchmarks is a poor predictor of success on financial tasks; (3) Traditional Bag-of-Words (BoW) models unexpectedly surpass all tested dense embedding models on financial STS tasks, highlighting persistent challenges for current embeddings in capturing nuanced financial semantics. 

Apart from these insights, our practical contributions are twofold: First, we propose FinMTEB, the first comprehensive financial domain evaluation benchmark encompassing 64 datasets across seven distinct tasks in both Chinese and English. Second, we develop and release Fin-E5, a finance-adapted embedding model that achieves state-of-the-art performance on FinMTEB. To support future research, we will make both the FinMTEB benchmark and our Fin-E5 model available as open source.

\section{Related Work}

Recent advances in embedding models have shown remarkable success in general domain tasks, yet their effectiveness in specialized domains remains a critical challenge. 

\subsection{General-purpose Embedding Models}
The evolution of embedding models marks significant progress in natural language processing. Starting with static word representations like Word2Vec \citep{word2vector} and GloVe \citep{pennington-etal-2014-glove}, the field advanced to contextualized embeddings through transformer-based architectures such as BERT \citep{Bert} and RoBERTa \citep{roberta}. A notable advancement came with Sentence-BERT \citep{sentence-bert}, which introduced Siamese and triplet network architectures to generate meaningful sentence-level representations. Recent developments in large language models have further pushed the boundaries, with models such as e5-mistral-7b-instruct \citep{e5} and gte-Qwen2-1.5B-instruct \citep{qwen2} achieving better performance in various embedding tasks. However, these general-purpose models may not adequately capture the nuanced semantics of specialized domains.

\subsection{Current Embedding Evaluation Landscape}
To assess embedding quality, several evaluation frameworks have been developed. General-purpose embedding benchmarks, such as the Massive Text Embedding Benchmark (MTEB) \citep{mteb}, provide broad coverage across multiple tasks and languages. Specialized benchmarks like BEIR \citep{beir} focus on specific aspects, such as information retrieval. Although they incorporate some domain-specific datasets, such as FiQA \citep{FiQA}, the size of the data and the coverage of the task are limited. 

\subsection{Domain Adaptation Approaches}
Recognizing the limitations of general-purpose models in specialized domains, researchers have pursued two main adaptation strategies. The first approach develops domain-specific models from scratch, exemplified by BioMedLM \citep{biomedlm} for biomedicine, SaulLM-7B \citep{saullm} for legal texts, and BloombergGPT \citep{bloomberggpt} for finance. The second strategy fine-tunes existing models for domain-specific tasks, as demonstrated by InvestLM \citep{investlm} and FinGPT \citep{fingpt}. This trend extends to embedding models, with specialized versions such as BioWordVec \citep{biowordvec}, BioSentVec \citep{biosentvec}, and FinBERT \citep{finbert} showing superior domain-specific performance. However, evaluating these specialized embedding models remains challenging due to the lack of comprehensive domain-specific benchmarks.

\subsection{The Gap in Domain-specific Evaluation}
While domain-specific language models have stimulated the development of specialized evaluation frameworks across various fields, these benchmarks primarily emphasize generative and reasoning capabilities instead of embedding quality. The financial sector has seen the emergence of frameworks like CFLUE \citep{CFLUE}, FinEval \citep{fineval}, and FinanceBench \citep{financebench}, whereas the legal and medical domains have introduced LawBench \citep{lawbench}, MedBench \citep{medbench}, and DrBenchmark \citep{drbenchmark}. These benchmarks consistently illustrate that general-purpose models often fall short in specialized areas \citep{CFLUE, lawbench}, highlighting the necessity of domain adaptation \citep{domainiskey}. Despite this acknowledgment, there is still a critical lack of comprehensive evaluation frameworks for domain-specific embeddings that assess performance across essential tasks such as semantic similarity, classification, and retrieval. Even recent financial embedding developments, such as BAM embedding~\citep{anderson-etal-2024-finance_text_embedding}, rely on narrow evaluation frameworks, typically focusing on single-task performance metrics (e.g., FinanceBench~\citep{financebench} for retrieval tasks). 
This limited evaluation may not fully reflect how the models perform in real-world financial applications.

\section{The FinMTEB Benchmark}

In this section, we introduce the Finance MTEB (FinMTEB) benchmark. As illustrated in Figure \ref{fig: overview}, FinMTEB encompasses seven embedding tasks, following a structure similar to MTEB \citep{mteb} but with datasets specifically curated for the finance domain.

\subsection{FinMTEB Tasks}

\textbf{Semantic Textual Similarity (STS)} evaluates the semantic similarity between pairs of financial text. This task is crucial for automated financial analysis and risk management; for example, detecting subtle semantic differences between quarterly earnings statements could reveal important shifts in a company's financial strategy that impact investment decisions. To ensure comprehensive evaluation, we incorporate diverse financial datasets, including FinSTS \citep{finsts} and FINAL \citep{final} from company annual reports, and BQ-Corpus \citep{bq-corpus} from banking documents. Model performance is quantified using Spearman's rank correlation, which measures the alignment between predicted cosine similarity scores and human-annotated similarity ratings.

\textbf{Retrieval} evaluates a model's capability to identify and extract relevant financial information in response to specific queries. Unlike general domain retrieval, financial information retrieval presents unique challenges, requiring precise handling of complex numerical data, temporal dependencies, and regulatory context. For comprehensive evaluation, we leverage established finance QA datasets including FinanceBench \citep{financebench}, FiQA2018 \citep{FiQA}, and HPC3 \citep{hpc3}. To further assess models' understanding of professional financial terminology, we introduce TheGoldman dataset, constructed from the Goldman Sachs Financial Dictionary. Performance is measured using NDCG@10, a metric that evaluates both the relevance of retrieved information and its ranking position, reflecting the real-world requirement for highly precise top results in financial applications.

\textbf{Clustering} evaluates a model's ability to automatically group similar financial texts based on their semantic content. To ensure comprehensive evaluation, we developed multiple specialized datasets that capture different aspects of financial text clustering: (1) FinanceArxiv-s2s and FinanceArxiv-p2p, constructed from titles and abstracts of finance-related papers on arXiv, providing rich academic financial content; (2) CompanyWiki2Industry dataset, derived from Wikipedia company descriptions, offering diverse industry categorization scenarios; and (3) complementary resources including consumer complaints from CFPB\footnote{https://huggingface.co/datasets/CFPB/consumer-finance-complaints}, financial intent detection data \citep{Intent_Detection,Synthetic}, and other established datasets. Model performance is quantified using the V-measure \citep{v-Measure}, a comprehensive metric that evaluates cluster quality through both completeness (all members of a class are assigned to the same cluster) and homogeneity (each cluster contains only members of a single class).

\textbf{Classification} evaluates a model's ability to categorize financial texts into predefined classes based on their semantic content. This capability is essential for automated financial decision-making; for example, in algorithmic trading, accurately classifying sentiment in earnings calls or news articles can directly influence trading strategies and portfolio adjustments. The classification task encompasses diverse financial scenarios through multiple specialized datasets, including: financial sentiment analysis \citep{fpb,FiQA,semeval, bbt}, Federal Reserve monetary policy classification \citep{fomc}, organization's strategy classification, and forward-looking statement identification \citep{investlm}. Performance is measured using Mean Average Precision (MAP), which provides a comprehensive assessment of classification accuracy while accounting for ranking quality and confidence scores.

\textbf{Reranking} evaluates the model's ability to order retrieved documents based on their relevance to financial queries. We utilize financial question-answering datasets such as Fin-Fact and FinQA\citep{rangapur2023finfact,finqa} to construct the reranking tasks. Specifically, for each query in these datasets, we retrieve top-k relevant documents along with the ground truth answers to construct the reranking training and evaluation pairs. The main evaluation metric for reranking in Finance MTEB is Mean Average Precision (MAP).

\textbf{Pair-Classification} evaluates a model's ability to determine semantic relationships between financial text pairs. This task includes two datasets: (1) the AFQMC dataset\footnote{https://tianchi.aliyun.com/dataset/106411} for customer intention, and (2) three financial news headline datasets \citep{headline}. We use Average Precision (AP) as the evaluation metric to assess model performance across different decision thresholds.

\textbf{Summarization} is evaluated based on how well the semantic similarity between an original text and its summary, as captured by embeddings, correlates with human judgments of summary quality. The evaluation corpus encompasses a comprehensive range of financial texts, including earnings call transcripts \citep{ectsum}, financial news articles \citep{bbt}, and SEC Form 10-K filings \citep{fns2022}, ensuring robust assessment across diverse financial contexts and writing styles.

\subsection{Characteristics of FinMTEB}
FinMTEB is constructed to provide a comprehensive evaluation platform for financial text embedding models. It encompasses a total of 64 datasets, specifically \textbf{35} datasets in English and \textbf{29} datasets in Chinese. Beyond the number of datasets, FinMTEB exhibits distinct linguistic and semantic properties crucial for domain-specific benchmarking. The descriptions of these individual datasets are available in Appendix~\ref{append: datasets}, and the detailed construction process for the new dataset is illustrated in the Appendix~\ref {app:dataset_construction}

\section{Fin-E5: Finance-Adapted Text Embedding Model}
Data is vital for domain adaptation~\citep{domainiskey}. However, existing public financial retrieval datasets exhibit a narrow scope, which creates a gap in training an LLM-based embedding model. For example, FiQA~\citep{FiQA}, a widely used financial retrieval dataset, primarily focuses on opinion-based content from online platforms, neglecting crucial aspects such as fundamental financial knowledge, technical terminology, and essential investment data. Thus, we start by curating a finance training dataset for adaptation.

\subsection{Data Formation}
We aim to construct each training instance as a triplet structure $(q, d^+, D^-)$, where $q$ represents a financial query, $d^+$ denotes a relevant document that provides substantive information addressing the query, and $D^-$ comprises carefully selected negative examples that share the financial domain but differ in semantic intent. 

\subsection{Training Data Construction}
To create a comprehensive dataset tailored for financial embedding training, we employ a systematic approach that combines expert-curated seed data with persona-based synthetic data generation.

\textbf{Seed Data. } Our seed data comes from the finance-specific QA dataset provided by InvestLM \citep{investlm}, which offers expert-validated financial content across various domains, such as market analysis, investment strategies, and corporate finance. To ensure evaluation integrity, we conduct rigorous overlap checks between our training data and the FinMTEB benchmark, guaranteeing no overlap.

\textbf{Persona-based Data Augmentation. } To enhance the diversity of financial task representations and generate varied (query, positive context, hard negative context) triplets for contrastive training, we develop a persona-based data augmentation framework derived from QA data generation~\citep{ge2024scaling}. Our framework employs a three-stage process that specifically targets the expansion of task coverage while preserving domain consistency: 

\begin{itemize}
    
    \item \textbf{Persona and Associated Task Identification:} We begin by analyzing each question-answer pair from our seed data. Using Qwen2.5-14B-Instruct~\citep{qwen2.5}
    with the prompt "\textit{Who is likely to use this text?}", the model generates a detailed persona description. This description inherently captures the persona (e.g., venture capitalist, financial advisor) and their typical job-related tasks (e.g., evaluating startup investments and managing client portfolios). For example, a generated description might be:
    \begin{tcolorbox}[
      colback=blue!5!white,  
      colframe=blue!60!black, 
      title=Example Persona\&Task Description, 
      fonttitle=\bfseries\sffamily\small, 
      breakable,               
      arc=2mm,                   
      boxsep=1mm,            
      left=2mm, right=2mm, top=2mm, bottom=2mm
    ]
    \textit{A \textbf{compliance officer} at a financial institution \textnormal{(Persona)}, responsible for \textbf{tracking major economic indicators and their potential regulatory implications} \textnormal{(Task)}, with a focus on market stability and accurate risk assessment.}
    \end{tcolorbox}
    

    \item \textbf{Contextual Query Generation:} Based on the rich persona description obtained in the previous step, we then prompt Qwen2.5-72B-Instruct~\citep{qwen2.5} to generate new queries $q$ that this persona might ask. The prompt used is: "\textit{Guess a prompt (i.e., instructions) that the following persona may ask you to do:}" The term "contextual" in this stage refers to our filtering process: we select queries that inherently require external documents or information for a comprehensive answer. This is crucial for forming the (query $q$, positive document $d^+$) pairs needed for training. For example, deriving from the compliance officer persona, the following example query would be considered contextual as it necessitates specific external analyses or regulatory interpretations:

    \begin{tcolorbox}[
      colback=blue!5!white,  
      colframe=blue!60!black, 
      title=Example Contextual Query $q$, 
      fonttitle=\bfseries\sffamily\small, 
      breakable,               
      arc=2mm,                   
      boxsep=1mm,            
      left=2mm, right=2mm, top=2mm, bottom=2mm
    ]
    \textit{What is the latest analysis on how the recent G7 central bank interest rate hikes might affect liquidity risk reporting for commercial banks?}
    \end{tcolorbox}
    
    \item \textbf{Synthetic Positive Document ($d^+$) Generation:} For each selected contextual query $q$, we synthesize a relevant positive financial document $d^+$. This document is generated using an LLM (e.g., Qwen2.5-72B-Instruct~\citep{qwen2.5}) with the prompt: "\textit{Synthesize context information related to this question: [Insert query $q$ here]}". The aim is for $d^+$ to provide substantive, focused information that directly addresses the query $q$, aligning with the information needs implied by the persona's role and their associated tasks. For the example query about EPS growth, the synthesized document would contain plausible (though synthetic) data, analyses, or relevant financial discussions.

    \item \textbf{Synthetic Positive Document ($d^+$) Generation:} For each selected contextual query $q$, we synthesize a relevant positive financial document $d^+$. This document is generated using an LLM (e.g., Qwen2.5-72B-Instruct~\citep{qwen2.5}) with the prompt: "\textit{Synthesize context information related to this question: [Insert query $q$ here]}". The aim is for $d^+$ to provide substantive, focused information that directly addresses the query $q$, aligning with the information needs implied by the persona's role and their associated tasks. For the example query about the impact of interest rate hikes on liquidity risk reporting, the synthesized document $d^+$ would contain plausible (though synthetic) expert analysis or excerpts from regulatory guidance, as illustrated below:
    \begin{tcolorbox}[
       colback=blue!5!white,  
      colframe=blue!60!black, 
      title=Example Synthesized Positive Document ($d^+$), 
      fonttitle=\bfseries\sffamily\small, 
      breakable,               
      arc=2mm,                   
      boxsep=1mm,            
      left=2mm, right=2mm, top=2mm, bottom=2mm
    ]
    \textit{A recent analysis by the Financial Monitoring Group, dated May 15, 2025, indicates that the coordinated interest rate increases by G7 central banks are anticipated to impact short-term funding markets significantly...}
    \end{tcolorbox}
    
\end{itemize}

\subsection{Training Pipeline}

Our primary objective in this training phase is to further adapt the e5-mistral-7b-instruct model~\citep{e5} to the financial domain's specific linguistic nuances and informational structures. This adaptation directly leverages the diverse financial query ($q$) and corresponding synthetic positive document ($d^+$) pairs generated through the persona-based data construction process detailed previously.

The foundation of our training methodology is a contrastive learning approach utilizing (query, positive context, hard negative context) triplets. Each training instance is structured as $(q, d^+, D^-)$, where:
\begin{itemize}
    \item $q$ represents the financial query, which serves as the anchor point for learning.
    \item $d^+$ is the synthetic document, specifically generated in our data construction phase to be a highly relevant positive contextual passage for the query $q$.
    \item $D^-$ denotes a set of hard negative contexts. These are documents also from the financial domain that, while potentially semantically similar to the query $q$ (making them challenging examples), are not the correct or directly relevant positive context $d^+$. To identify these hard negatives, we employ an auxiliary embedding model, all-MiniLM-L12-v2~\citep{sentence-bert}, to mine for documents that are close to $q$ in its embedding space but are distinct from $d^+$.
\end{itemize}

In line with the training recipe for e5-mistral-7b-instruct~\citep{e5}, we utilize the last token pooling method to derive fixed-size embeddings for both queries and documents. The e5-mistral-7b-instruct model is then fine-tuned using these $(q, d^+, D^-)$ triplets. The training process is guided by the InfoNCE (Noise Contrastive Estimation) loss function~\citep{oord2018infonce}. This loss function incentivizes the model to learn representations where the embedding of the query $q$ is closer to the embedding of its positive context $d^+$ compared to its distance from the embeddings of all hard negative contexts $D^-$ within the same training batch (referred to as in-batch negatives).

Full details regarding the fine-tuning process, including specific hyperparameters (such as batch sizes and learning rates), any input formatting templates utilized, and optimization settings for adapting e5-mistral-7b-instruct, are comprehensively documented in Appendix~\ref{append: e5}.






\section{Experimental Evaluation}

In this section, we conduct a comprehensive evaluation of various embedding models on FinMTEB. Our primary goals are to benchmark their performance in the financial domain, analyze the impact of different model characteristics (such as domain adaptation and architecture), and investigate the necessity of domain-specific benchmarks like FinMTEB. Since most of the evaluated pre-trained models are predominantly trained on English corpora, our main evaluation focuses on the English datasets within FinMTEB; the evaluation results based on Chinese datasets are illustrated in Appendix \ref{app:zh}. The benchmark time is reported in Appendix \ref{append: time}. 

\begin{center}
   \begin{table*}[htbp]
\centering
\resizebox{\linewidth}{!}{%
\begin{tabular}{llccccccccc}
\toprule
\multirow{3}{*}{\textbf{Model}} & \multirow{3}{*}{\textbf{Size}} & \multicolumn{7}{c}{\textbf{Tasks}} & \multirow{3}{*}{\textbf{Avg.}} \\
\cmidrule(lr){3-9}  
& & \textbf{STS} & \textbf{Retrieval} & \textbf{Class.} & \textbf{Cluster.} & \textbf{Rerank.} & \textbf{PairClass.} & \textbf{Summ.} & \\
\cmidrule(lr){3-9}
& & \footnotesize{(N=\textbf{2}, p=0.10)} 
  & \footnotesize{(N=\textbf{10}, p$<$0.05\textsuperscript{*})} 
  & \footnotesize{(N=\textbf{8}, p$<$0.05\textsuperscript{*})} 
  & \footnotesize{(N=\textbf{6}, p=0.12)} 
  & \footnotesize{(N=\textbf{3}, p$<$0.05\textsuperscript{*})} 
  & \footnotesize{(N=\textbf{3}, p$<$0.05\textsuperscript{*})} 
  & \footnotesize{(N=\textbf{3}, p=0.45)} & \\ 
\midrule
BOW & - &\textbf{0.4845} & 0.2084 & 0.4696 & 0.2547 & 0.7628 & 0.7143 & 0.2584 & 0.4504 \\
\hline
\multicolumn{10}{l}{\textbf{Encoder based Models}} \\ 
\hline
BERT & 110M  & 0.3789 & 0.0207 & 0.5496 & 0.1744 & 0.3930 & 0.7111 & 0.1686 & 0.3423 \\
FinBERT & 110M & 0.4198 & 0.1102 & 0.5923 & 0.2833 & 0.6404 & 0.6967 & 0.2010 & 0.4205 \\
instructor-base & 110M & 0.3732 & 0.5772 & 0.6208 & 0.5300 & 0.9734 & 0.6138 & 0.4315 & 0.5886 \\
bge-large-en-v1.5 & 335M & 0.3396 & 0.6463 & 0.6436 & 0.5725 & 0.9825 & 0.7400 & 0.4857 & 0.6301 \\
AnglE-BERT & 335M & 0.3080 & 0.5730 & 0.6439 & 0.5774 & 0.9650 & 0.6891 & 0.5049 & 0.6088 
\\
\hline
\multicolumn{10}{l}{\textbf{LLM-based Models}} \\ 
\hline
gte-Qwen1.5-7B-instruct & 7B & 0.3758 & 0.6697 & 0.6438 & 0.5854 & 0.9890 & 0.6998 & 0.5354 & 0.6427 \\
Echo & 7B & \underline{0.4380} & 0.6443 & 0.6525& 0.5776 & 0.9765 & 0.6261 & 0.4722 & 0.6267 \\
bge-en-icl & 7B & 0.3233 & 0.6789 & 0.6569 & 0.5742 & 0.9898 & 0.6738 & 0.5197 & 0.6309 \\
NV-Embed v2 & 7B & 0.3739 & 0.7061 & 0.6393 & 0.6096 & 0.9822 & 0.6043 & 0.5103 & 0.6322 \\
e5-mistral-7b-instruct & 7B & 0.3800 & 0.6749 & 0.6449 & 0.5783 & 0.9875 & \underline{0.7394} & 0.5275 & 0.6475 \\
\hline
\multicolumn{10}{l}{\textbf{Commercial Models}} \\ 
\hline
text-embedding-3-small & - & 0.3254 & 0.6641 & 0.6387 & 0.5802 & 0.9825 & 0.5957 & 0.5085 & 0.6136 \\
text-embedding-3-large & - & 0.3615 & \underline{0.7112} & 0.6596 & \textbf{0.6081} & \underline{0.9910} & 0.7309 & \underline{0.5671} & 0.6613 \\
voyage-3-large & - & 0.4145 & \textbf{0.7463} & \underline{0.6861} & \underline{0.5944} & \textbf{0.9938} & 0.6519 & \textbf{0.6484} & \underline{0.6765}\\
\hline
\multicolumn{10}{l}{\textbf{Finance Adapted LLM-based Models}} \\ 
\hline
Fin-E5 & 7B & 0.4342 & 0.7105 & \textbf{0.7565\textsuperscript{†}} & 0.5650 & 0.9896 & \textbf{0.8014} & 0.4797 & \textbf{0.6767} \\
\bottomrule
\end{tabular}%
}
\caption{Performance comparison across different embedding models on FinMTEB benchmark. Tasks include semantic textual similarity (STS), retrieval, classification (Class.), clustering (Cluster.), reranking (Rerank.), pair classification (PairClass.), and summarization (Summ.). For each task, N indicates the number of datasets, and p is the p-value from one-way ANOVA testing for significant differences across all models; * denotes $p < 0.05$. \textbf{Bold} indicates best performance, \underline{underline} indicates second-best, and † indicates statistically significant differences between the best two models (p < 0.05).}
\label{tab:benchmark}
\end{table*}
 
\end{center}

\subsection{Experimental Setup}
\label{subsec:exp_setup}

\textbf{Evaluated Models}
In addition to Fin-E5, our proposed finance-adapted model, we evaluate four broad categories of existing embedding models on the FinMTEB benchmark. These include:

\begin{itemize}
\item \textbf{Bag-of-Words (BOW):} A traditional baseline representing text as sparse vectors based on word frequencies.

\item \textbf{Encoder-based Models:} This category includes various transformer encoder architectures: (1) classical models like BERT (CLS pooling)~\citep{Bert} and the domain-specific FinBERT~\citep{finbert}; (2) models optimized for semantic search such as msmarco-bert-base-dot-v5 and all-MiniLM-L12-v2~\citep{sentence-bert}; and (3) advanced architectures including bge-large-en-v1.5~\citep{bge_embedding}, AnglE-BERT~\citep{li2023angle}, and instructor-base~\citep{instructor}.

\item \textbf{LLM-based Models:} We investigate several state-of-the-art decoder-based or LLM-enhanced embedding models: (1) Mistral-7B-based models including bge-en-icl (Mistral-7B backbone with further instruction tuning)~\citep{bge_embedding}, e5-mistral-7b-instruct~\citep{e5}, and Echo~\citep{echo}; (2) NV-Embed v2~\citep{NV-Embed}; and (3) gte-Qwen1.5-7B-instruct~\citep{gte}, built on the Qwen~\citep{qwen2} architecture.

\item \textbf{Commercial Models:} For a comprehensive comparison, we include leading closed-source commercial solutions, specifically OpenAI's text-embedding-3-large, text-embedding-3-small~\citep{openai_embedding}, and voyage-3-large~\citep{voyage}\footnote{We thank Voyage AI for providing API credits that supported us in conducting the evaluation with their model.}. 

\end{itemize}

\subsection{Overall Performance on FinMTEB}

The comprehensive performance of all evaluated models across the various tasks in the FinMTEB benchmark is presented in Table~\ref{tab:benchmark}. This table serves as the primary basis for the subsequent analyses.

\subsubsection{Impact of Domain Adaptation}
\label{subsubsec:domain_adaptation}

Domain specialization considerably boosts performance on financial tasks, as illustrated in Table~\ref{tab:benchmark}. For instance, the finance-specific FinBERT outperforms the general BERT by 15.6\% in the average score (0.6721 vs. 0.5812 on relevant FinMTEB tasks). Similarly, our finance-adapted Fin-E5 model exceeds its general-domain counterpart, e5-mistral-7b-instruct, by 4.5\% in the average score. This overall improvement is supported by statistically significant gains in several key task categories, as detailed in Table~\ref{tab:fin_e5_vs_baseline_significance}. Specifically, Fin-E5 demonstrates a significant advantage in Classification, achieving a score of 0.7565 compared to the baseline's 0.6449 (p = 0.0206), and also in Retrieval, scoring 0.7105 against the baseline's 0.6749 (p = 0.0489). Fin-E5’s slight underperformance on Clustering and Summarization compared with e5-mistral-7b-instruct is not statistically significant (p > 0.05).
Fin-E5 also achieves state-of-the-art performance (0.6767 average scores) on FinMTEB, surpassing general-purpose, open-source, and leading commercial models. This increased performance comes from an efficient adaptation process requiring only 100 training steps.

\subsubsection{Limitations of Current Models in Financial STS Tasks}

The Semantic Textual Similarity (STS) task results reveal a counterintuitive finding: the simple BOW model (achieving a score of 0.4845) outperforms all evaluated dense embedding architectures on STS. The observation highlights fundamental limitations in dense embedding strategies for specialized financial documents. The STS datasets \cite{finsts,final} are sourced from the Company Annual Reports. 
Thus, this reversal of typical performance hierarchies likely arises from the specialized financial corpus, which can decrease performance for models not finely tuned to this vocabulary, whereas BOW benefits from exact term matches in such standardized disclosures. 

\section{The Necessity of Domain-Specific Benchmarks: An ANOVA Study}

This section addresses another research question. \textit{To what extent do general-purpose embedding evaluations appropriately capture domain-specific performance?} To investigate this, we conduct a quantitative comparison between the general-purpose MTEB benchmark~\citep{mteb} and our domain-specific FinMTEB. We employ Analysis of Variance to examine the main effects of two key factors, the embedding model (Model Factor) and the benchmark domain (Domain Factor: General vs. Finance), on model performance. Detailed experimental settings are provided in Appendix \ref{appendix: domain-bench}. The results reveal that the Domain Factor demonstrates statistical significance across all tasks (p < 0.001), with large F statistics in classification, clustering, and STS. These findings indicate that domain-specific characteristics significantly influence embedding model evaluation. 


\section{Conclusion}
This paper introduces FinMTEB, the first comprehensive benchmark for evaluating embedding models in the financial domain. Our main contributions include establishing a large-scale evaluation framework with 64 datasets across seven tasks in Chinese and English, and developing Fin-E5, a finance-adapted embedding model demonstrating competitive performance through persona-based data augmentation. Our empirical results highlight the importance of domain-specific adaptation and reveal current limitations in financial text embeddings. We believe FinMTEB will serve as a valuable resource for both researchers and practitioners in advancing financial language models.

\section*{Limitation}
This work has two primary limitations. First, it relies on several existing financial datasets that could potentially overlap with the training data of contemporary embedding models. This overlap may introduce contamination, making it difficult to ensure completely fair comparisons between different models. Second, our adapted model and evaluation methods are currently limited to the English language, which restricts their applicability to non-English financial texts.

\bibliography{acl_latex}

\appendix


\section{Datasets in FinMTEB}
\label{append: datasets}
The detailed description of each dataset used in this work is listed in the Table \cref{tab:class_ds,tab:rerank_ds,tab:cluster_ds,tab:sum_ds,tab:sts_ds,tab:PairClassification_ds,tab:retrieval_ds}.

\begin{table*}[htbp]
\resizebox{\textwidth}{!}{
\centering
\begin{tabular}{llp{9cm}}
\toprule
\textbf{Dataset Name} & \textbf{Language} & \textbf{Description} \\
\toprule
FINAL \citep{final} & English & A dataset designed for discovering financial signals in narrative financial reports. \\

FinSTS \citep{finsts} & English & A dataset focused on detecting subtle semantic shifts in financial narratives. \\

AFQMC \tablefootnote{\url{https://tianchi.aliyun.com/dataset/106411}} & Chinese & A Chinese dataset for customer service question matching in the financial domain. \\

BQ-Corpus \citep{bq-corpus} & Chinese & A large-scale Chinese corpus for sentence semantic equivalence identification (SSEI) in the banking domain. \\
\bottomrule
\end{tabular}

}
\caption{Summary of STS Datasets}
\label{tab:sts_ds}
\end{table*}

\begin{table*}[htbp]
\resizebox{\textwidth}{!}{
\centering
\begin{tabular}{p{5cm}p{1.5cm}p{8cm}}
\toprule
\textbf{Dataset Name} & \textbf{Language} & \textbf{Description} \\
\toprule
FiQA2018 \citep{FiQA} & English & Financial opinion mining and question answering dataset. \\

FinanceBench \citep{financebench} & English & Open book financial question answering dataset. \\

HC3(Finance) \citep{hpc3} & English & A human-ChatGPT comparison corpus in the finance domain. \\

Apple-10K-2022 \tablefootnote{\url{https://lighthouz.ai/blog/rag-benchmark-finance-apple-10K-2022/}} & English & A retrieval-augmented generation (RAG) benchmark for finance applications. \\

FinQA \citep{finqa}& English & Financial numerical reasoning dataset with structured and unstructured evidence. \\

TAT-QA \citep{tatqa}& English & Question answering benchmark combining tabular and textual content in finance. \\

US Financial News \tablefootnote{\url{https://www.kaggle.com/datasets/jeet2016/us-financial-news-articles}}& English & Finance news articles paired with headlines and stock ticker symbols. \\

TradeTheEvent (Trading Benchmark) \citep{trade} & English & Finance news articles paired with headlines and stock ticker symbols. \\

TradeTheEvent (Domain Adaption) \citep{trade} & English & Financial terms and explanations dataset. \\

TheGoldman-en & English & English version of the Goldman Sachs Financial Dictionary. \\

FinTruthQA \citep{fintruthqa} & Chinese& Dataset for evaluating the quality of financial information disclosure. \\

Fin-Eva (Retrieval task) \tablefootnote{\url{https://github.com/alipay/financial_evaluation_dataset/tree/main}} & Chinese& Financial scenario QA dataset focusing on retrieval tasks. \\

AlphaFin \citep{alphafin}& Chinese& Comprehensive financial dataset including NLI, QA, and stock trend predictions. \\

DISC-FinLLM (Retrieval Part Data) \citep{disc}& Chinese& Financial scenario QA dataset. \\

FinQA (from DuEE-fin) \citep{bbt}& Chinese& Financial news bulletin event quiz dataset. \\

DISC-FinLLM (Computing) \citep{disc}& Chinese& Financial scenario QA dataset focusing on numerical tasks. \\

SmoothNLP \tablefootnote{\url{https://github.com/smoothnlp/SmoothNLP}} & Chinese& Chinese finance news dataset. \\

THUCNews \citep{thuctc}& Chinese& Chinese finance news dataset. \\

Fin-Eva (Terminology) \tablefootnote{\url{https://github.com/alipay/financial_evaluation_dataset/tree/main}} & Chinese& Financial terminology dataset used in the industry. \\
TheGoldman-cn & Chinese& Chinese version of the Goldman Sachs Financial Dictionary. \\
\bottomrule
\end{tabular}
}
\caption{Summary of Retrieval Datasets}
\label{tab:retrieval_ds}
\end{table*}

\begin{table*}[htbp]
\resizebox{\textwidth}{!}{
\centering
\begin{tabular}{llp{9cm}}
\toprule
\textbf{Dataset Name} & \textbf{Language} & \textbf{Description} \\
\toprule
FinancialPhrasebank \citep{fpb} & English & Polar sentiment dataset of sentences from financial news, categorized by sentiment into positive, negative, or neutral. \\ 
FinSent \citep{investlm} & English & Polar sentiment dataset of sentences from the financial domain, categorized by sentiment into positive, negative, or neutral. \\ 
FiQA\_ABSA \citep{FiQA} & English & Polar sentiment dataset of sentences from the financial domain, categorized by sentiment into positive, negative, or neutral. \\ 
SemEva2017\_Headline \citep{semeval}& English & Polar sentiment dataset of sentences from the financial domain, categorized by sentiment into positive, negative, or neutral. \\ 
FLS \citep{investlm}& English & A finance dataset detects whether the sentence is a forward-looking statement. \\ 
ESG \citep{investlm}& English & A finance dataset performs sentence classification under the environmental, social, and corporate governance (ESG) framework. \\ 
FOMC \citep{fomc}& English & A task of hawkish-dovish classification in finance domain. \\ 
Financial-Fraud \tablefootnote{\url{https://github.com/amitkedia007/Financial-Fraud-Detection-Using-LLMs/tree/main}}& English & This dataset was used for research in detecting financial fraud. \\ 
FinNSP \citep{bbt} & Chinese & Financial negative news and its subject determination dataset. \\ 
FinChina \citep{finchinese}& Chinese & Polar sentiment dataset of sentences from the financial domain, categorized by sentiment into positive, negative, or neutral. \\ 
FinFE \citep{bbt}& Chinese & Financial social media text sentiment categorization dataset. \\ 
OpenFinData \tablefootnote{\url{https://github.com/open-compass/OpenFinData?tab=readme-ov-file}} & Chinese & Financial scenario QA dataset including sentiment task. \\ 
MDFEND-Weibo2 (finance) \citep{nan2021mdfend} & Chinese & Fake news detection in the finance domain. \\ 
\bottomrule
\end{tabular}
}
\caption{Summary of Classification Datasets}
\label{tab:class_ds}
\end{table*}

\begin{table*}[htbp]
\resizebox{\textwidth}{!}{
\centering
\begin{tabular}{llp{9cm}}
\toprule
\textbf{Dataset Name} & \textbf{Language} & \textbf{Description} \\
\toprule
MInDS-14-en \citep{MInDS}& English & MINDS-14 is a dataset for intent detection in e-banking, covering 14 intents across 14 languages. \\ 
Consumer Complaints \citep{cfpb}& English & The Consumer Complaint Database is a collection of complaints about consumer financial products and services that sent to companies for response. \\ 
Synthetic PII finance \citep{Synthetic} & English & Synthetic financial documents containing Personally Identifiable Information (PII). \\ 
FinanceArxiv-s2s & English & Clustering of titles from arxiv (q-fin). \\ 
FinanceArxiv-p2p & English & Clustering of abstract from arxiv (q-fin). \\ 
WikiCompany2Industry-en & English & Clustering the related industry domain according to the company description. \\ 
MInDS-14-zh \citep{MInDS}& Chinese & MINDS-14 is a dataset for intent detection in e-banking, covering 14 intents across 14 languages. \\ 
FinNL \citep{bbt}& Chinese & Financial news categorization dataset. \\ 
CCKS2022 \citep{CCKS}& Chinese & Clustering of financial events. \\ 
CCKS2020 \citep{CCKS}& Chinese & Clustering of financial events. \\ 
CCKS2019 \citep{CCKS}& Chinese & Clustering of financial events. \\
\bottomrule
\end{tabular}

}
\caption{Summary of Clustering Datasets}
\label{tab:cluster_ds}
\end{table*}

\begin{table*}[htbp]

\resizebox{\textwidth}{!}{
\centering
\begin{tabular}{llp{9cm}}
\toprule
\textbf{Dataset Name} & \textbf{Language} & \textbf{Description} \\
\toprule
Ectsum \citep{ectsum}& English & A Dataset For Bullet Point Summarization of Long Earnings Call Transcripts.  \\ 
FINDSum \citep{findsum}& English & A Large-Scale Dataset for Long Text and Multi-Table Summarization.  \\ 
FNS-2022 \citep{fns2022}& English & Financial Narrative Summarisation for 10K.  \\ 
FiNNA \citep{bbt}& Chinese & A financial news summarization dataset.  \\ 
Fin-Eva (Headline) \citep{fineval} & Chinese & A financial summarization dataset.   \\ 
Fin-Eva (Abstract) \citep{fineval}& Chinese & A financial summarization dataset.  \\
\bottomrule
\end{tabular}
}
\caption{Summary of Summarization Datasets}
\label{tab:sum_ds}
\end{table*}

\begin{table*}[htbp]

\resizebox{\textwidth}{!}{
\centering
\begin{tabular}{llp{9cm}}
\toprule
\textbf{Dataset Name} & \textbf{Language} & \textbf{Description} \\
\toprule
Fin-Fact \citep{rangapur2023finfact}& English & A Benchmark Dataset for Financial Fact Checking and Explanation Generation. \\
FiQA2018 \citep{FiQA}& English & Financial opinion mining and question answering. \\
HC3(Finance) \citep{hpc3}& English & A human-ChatGPT comparison finance corpus. \\
Fin-Eva (Retrieval task) \citep{fineval}& Chinese & Financial scenario QA dataset including retrieval task. \\
DISC-FinLLM (Retrieval Part Data) \citep{disc}& Chinese & Financial scenario QA dataset. \\
\bottomrule
\end{tabular}
}

\caption{Summary of Reranking Datasets}
\label{tab:rerank_ds}
\end{table*}

\begin{table*}[htbp]

\resizebox{\textwidth}{!}{
\centering

\begin{tabular}{llp{9cm}}
\toprule
\textbf{Dataset Name} & \textbf{Language} & \textbf{Description} \\
\toprule
HeadlineAC-PairClassification \citep{headline}& English & Financial text sentiment categorization dataset.  \\ 
HeadlinePDD-PairClassification \citep{headline}& English & Financial text sentiment categorization dataset.  \\ 
HeadlinePDU-PairClassification \citep{headline}& English & Financial text sentiment categorization dataset.  \\ 
AFQMC & Chinese & Ant Financial Question Matching Corpus.  \\ 
\bottomrule
\end{tabular}
}

\caption{Summary of PairClassification Datasets}
\label{tab:PairClassification_ds}
\end{table*}

\subsection{Detailed Characteristics of FinMTEB}

\label{app:finmteb_characteristics_details} 

\textbf{Linguistic Pattern.} Table \ref{tab:compare_benchmark} presents a comparative analysis of linguistic features between MTEB~\citep{mteb} and FinMTEB benchmarks, examining aspects such as average sentence length, token length, syllables per token, and dependency distance \citep{dependency_distance}. The results indicate that texts in FinMTEB consistently exhibit longer and more complex sentences than those in MTEB, with an average sentence length of 26.37 tokens compared to MTEB's 18.2 tokens. This highlights the linguistic differences between financial and general domain texts.

\begin{center}
    \begin{figure*}[ht]
    \centering
    \includegraphics[width=\textwidth]{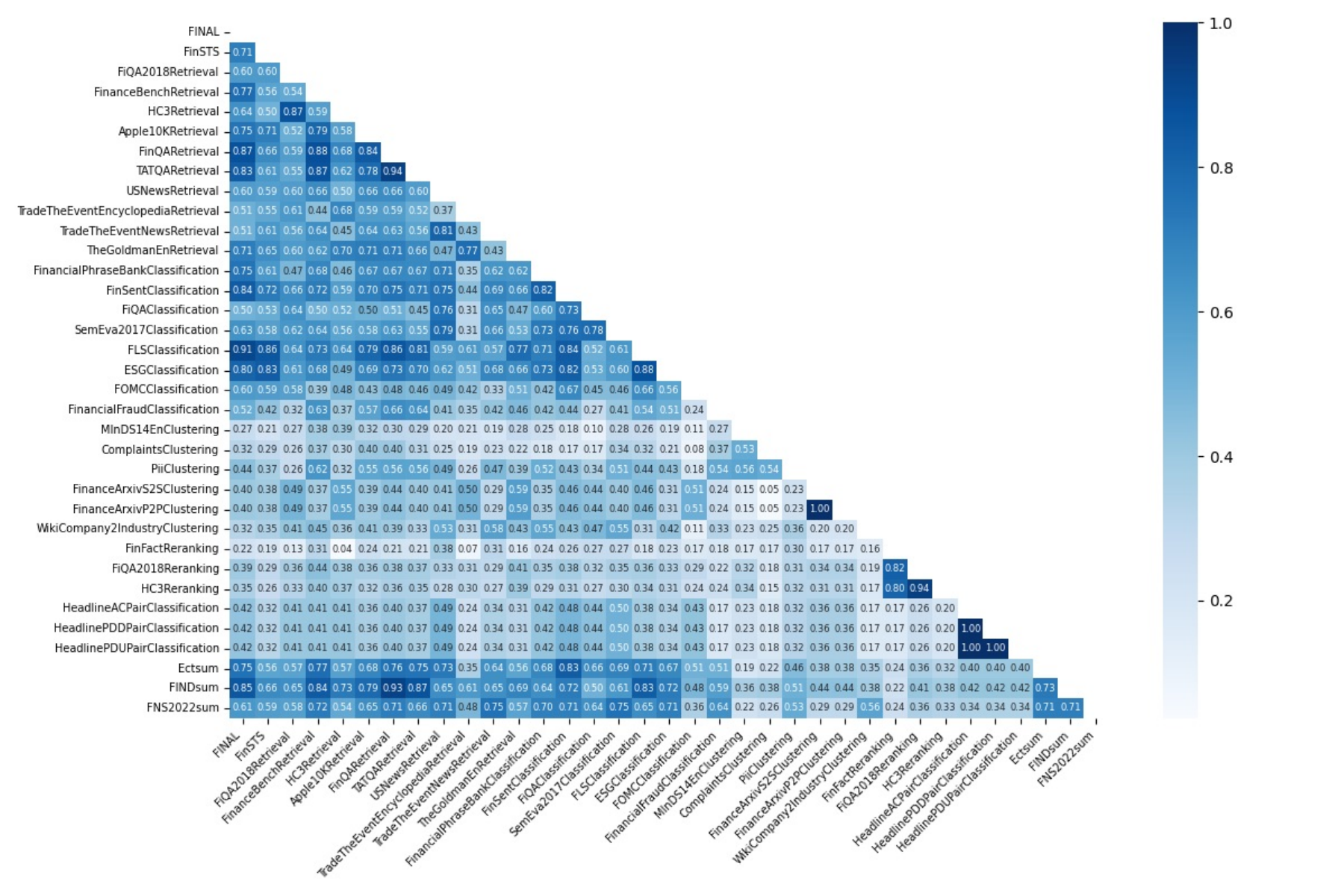}
    \caption{Semantic similarity across all the datasets in FinMTEB benchmark.}
    \label{fig: semantic_diveristy}
    \label{append: semantic_diveristy}
    \end{figure*} 
\end{center}

\textbf{Semantic Diversity.} We examine the inter-dataset semantic similarity within FinMTEB. Using the all-MiniLM-L6-v2 model\footnote{https://huggingface.co/sentence-transformers/all-MiniLM-L6-v2}, we embed 1,000 randomly sampled texts from each dataset, compute their mean embeddings to represent each dataset, and measure inter-dataset similarities using cosine similarity. As shown in Figure \ref{fig: semantic_diveristy}, most datasets in FinMTEB display inter-dataset similarity scores below 0.6, with a mean cosine similarity of 0.4, indicating semantic distinctions among various types of financial texts.

\begin{center}
    \begin{table*}[htb]
    \centering
    \begin{adjustbox}{width=0.8\linewidth}
    \begin{tabular}{ccccc}
        \toprule 
        \textbf{Benchmark} & \textbf{Sentence Length} & \textbf{Token Length}  & \textbf{Syllables Per Token} & \textbf{Dependency Distance}  \\
        \midrule
         MTEB & 18.20 & 4.89 & 1.49 & 2.49\\
        FinMTEB & 26.37 & 5.12 & 1.52 & 2.85\\
        \bottomrule
    \end{tabular}
    \end{adjustbox}
    \caption{Comparison of Text Characteristics Between FinMTEB and MTEB. The numbers represent the average scores across all samples from all datasets.}
    \label{tab:compare_benchmark}
    \end{table*}
\end{center}

\subsection{Dataset Construction and Validation Details}
\label{app:dataset_construction}

This appendix details the construction pipelines and validation procedures for all newly created datasets used in our experiments. Each dataset is carefully curated to ensure high quality and appropriate task alignment.

\textbf{WikiCompany2Industry (Clustering Task).} The WikiCompany2Industry dataset is constructed by extracting 3,820 company descriptions from Wikipedia's company pages and cross-referencing them with Standard Industrial Classification (SIC) codes from the SEC EDGAR database. To ensure data quality, we apply a three-stage filtering process that removes companies with multiple primary SIC codes, excludes descriptions shorter than 50 words, and verifies SIC code consistency across multiple data sources. The dataset uses official SIC 4-digit codes as clustering labels, resulting in 87 unique industry categories that provide comprehensive coverage of the business landscape.

\textbf{FinanceArxiv-s2s/p2p (Clustering Task).} The FinanceArxiv dataset is created by collecting all finance papers (titles and abstracts) from ArXiv spanning 2015-2024 using the official ArXiv API. The dataset includes two variants: s2s (sentence-to-sentence) for title-based clustering and p2p (paper-to-paper) for abstract-based clustering. The clustering is based on ArXiv's 9 official finance subcategories: q-fin.CP (Computational Finance), q-fin.EC (Economics), q-fin.GN (General Finance), q-fin.MF (Mathematical Finance), q-fin.PM (Portfolio Management), q-fin.PR (Pricing of Securities), q-fin.RM (Risk Management), q-fin.ST (Statistical Finance), and q-fin.TR (Trading and Market Microstructure).

\textbf{TheGoldman (Retrieval Task).} TheGoldman dataset was constructed by transforming 1,500 term-definition pairs from the Goldman Sachs Financial Dictionary into a query-retrieval format. Each query follows the standardized format "What is [term]?" with the corresponding definition serving as the target retrieval result. For example, the query "What is IPO Lock-up?" maps to the target "A legally binding contract between underwriters and insiders that prohibits the sale of shares for a specified period after an initial public offering..." This format ensures that the retrieval task evaluates the model's ability to understand financial terminology and provide accurate, contextually appropriate definitions.

\section{Training Details For Fin-E5}\label{append: e5}
The training dataset size is 19,467. The model is trained for 100 steps using the augmented dataset with a batch size of 128. For optimization, we use the AdamW optimizer with a learning rate of 1e-5 and implement a linear warmup schedule.  For a given data $(q, d^+, D^-)$, we adopt an instruction-based methodology for embedding training. The instruction template is as follows:

\begin{equation}
    q_{\text{inst}} = \text{Instruct: } \{task\_definition\} \textbackslash n \{q\}
\end{equation}

where $\{task\_definition\}$ represents a concise single-sentence description of the embedding task. 

\section{Chinese Dataset Evaluation in FinMTEB}
\label{app:zh}
Table \ref{tab:zh_results} presents the different performances of the model in Chinese evaluation datasets.
\begin{table*}[htbp]
\resizebox{\textwidth}{!}{
\centering
\begin{tabular}{@{}lcccccccc@{}}
\toprule
Model & STS & Retrieval & Class. & Cluster. & Rerank. & Pair-Class. & Summ. & Avg. \\
\midrule
BOW                                  & 0.2030 & 0.3000 & 0.4694 & 0.4204 & 0.9089 & 0.3376 & 0.3433 & 0.4260 \\
all-MiniLM-L12-v2                    & 0.1454 & 0.1777 & 0.4398 & 0.2243 & 0.7943 & 0.3375 & 0.4731 & 0.3703 \\
paraphrase-multilingual-MiniLM-L12-v2 & 0.2775 & 0.3795 & 0.5587 & 0.4612 & 0.9673 & 0.3882 & 0.3442 & 0.4824 \\
bge-large-zh-v1.5                    & 0.5806 & 0.6073 & 0.5996 & 0.6672 & 0.9931 & 0.5506 & 0.4413 & 0.6342 \\
bge-m3                               & 0.5083 & 0.6243 & 0.6209 & \textbf{0.7109} & 0.9902 & 0.5331 & 0.3582 & 0.6208 \\
multilingual-e5-large-instruct       & 0.4799 & 0.6303 & 0.5908 & 0.6540 & 0.9876 & 0.4651 & 0.4456 & 0.6076 \\
gte-Qwen1.5-7B-instruct              & \textbf{0.5714} & 0.6420 & 0.6200 & 0.6172 & 0.9921 & \textbf{0.5968} & \textbf{0.4934} & 0.6475 \\
text-embedding-3-large               & 0.3848 & 0.6778 & 0.6041 & 0.7054 & \textbf{1.0000} & 0.4547 & 0.4203 & 0.6067 \\
Fin-E5                               & 0.4799 & \textbf{0.6893} & \textbf{0.6681} & 0.6737 & 0.9931 & 0.5303 & 0.4207 & \textbf{0.6364} \\
\bottomrule
\end{tabular}

}
\caption{Performance comparison across Chinese datasets. This evaluation contains some multilingual models and Fin-E5.  The evaluation metrics include semantic textual similarity (STS), retrieval, classification (Class.), clustering (Cluster.), reranking (Rerank.), pair classification (PairClass.), and summarization (Summ.). \textbf{Best} results are in bold. }
\label{tab:zh_results}
\end{table*}

\section{Benchmarking Time Reporting.} \label{append: time}
The benchmarking was conducted on the NVIDIA H800 GPU using a batch size of 512. Echo Embedding~\citep{echo} required the longest processing time at 12 hours, followed by BeLLM~\citep{li2023angle} at 11.98 hours. AnglE-BERT~\citep{li2023angle} completed the evaluation in 8 hours, while NV-Embed v2~\citep{NV-Embed} demonstrated the highest efficiency, completing all tasks in just 5.6 hours. 

\section{Domain-specific Embedding Benchmark is needed}
\label{appendix: domain-bench}
This section addresses another research question. \textit{To what extent do general-purpose embedding evaluations appropriately capture domain-specific performance?} 
To solve this question, we run a quantitative comparison between MTEB~\citep{mteb} and FinMTEB.

\textbf{Models.} We evaluate \textbf{seven} state-of-the-art general-purpose embedding model. Specifically, we consider the following models: bge-en-icl \citep{bge_embedding} and e5-mistral-7b-instruct \citep{e5}, which are developed from Mistral-7B-v0.1 \citep{mistral}; gte-Qwen2-1.5B-instruct \citep{gte}, developed from Qwen2 \citep{qwen2}; bge-large-en-v1.5 \citep{bge_embedding} and all-MiniLM-L12-v2 \citep{sentence-bert}, both developed from BERT \citep{Bert}; instructor-base \citep{instructor} from T5Encoder \citep{t5Encoder}; and OpenAI’s text-embedding-3-small \citep{openai_embedding}. The overall score for these models in MTEB \citep{mteb} and FinMTEB is shown in Table \ref{tab:model_specs_and_scores}. 

\begin{table*}[htbp]
\centering
\begin{tabular}{@{}llrrr@{}}
\toprule
Embedding Model & Base Model & Dimensions & MTEB Score & FinMTEB Score \\
\midrule
bge-en-icl & Mistral & 4096 & 71.67 & 63.09 \\
gte-Qwen2-1.5B-instruct & Qwen2 & 1536 & 67.16 & 59.98 \\
e5-mistral-7b-instruct & Mistral & 4096 & 66.63 & 64.75 \\
bge-large-en-v1.5 & Bert & 1024 & 64.23 & 58.95 \\
text-embedding-3-small & --- & 1536 & 62.26 & 61.36 \\
instructor-base & T5Encoder & 768 & 59.54 & 54.79 \\
all-MiniLM-L12-v2 & Bert & 384 & 56.53 & 54.31 \\
\bottomrule
\end{tabular}
\caption{Comparison of Various Embedding Models: Performance on MTEB and FinMTEB Benchmarks}
\label{tab:model_specs_and_scores}
\end{table*}

\textbf{Method.} To ensure robust statistical analysis, we use bootstrapping methods to generate a large sample dataset. For each task in both MTEB and FinMTEB, we aggregate the datasets associated with the task into a task pool. From each task pool, we randomly select 50 examples to create a bootstrap sample and evaluate the embedding model’s performance on this bootstrap. We repeat this process 500 times, resulting in 500 bootstraps for each combination. Thus, we have 14 unique combinations (model and domain), each with 500 bootstraps and their corresponding performance scores.

\textbf{Analysis of Variance.} We conduct an Analysis of Variance (ANOVA) that examines the effects of both the model and the domain. The results reveal that the Domain Factor demonstrates statistical significance across all tasks (p < 0.001), with notably large F statistics in classification (F = 2086.30), clustering (F = 32161.37), and STS (F = 25761.71). Furthermore, the Domain Factor generally accounts for a greater share of the variance than the Model Factor, as indicated by the Sum of Squares (e.g., in Classification: Domain = 56.82 vs. Model = 4.17). These findings suggest that domain-specific characteristics significantly impact model performance, reinforcing the importance of specialized evaluation frameworks such as FinMTEB for financial applications.

\section{Spearman’s Correlation of Embedding Models’ Performance}
We evaluate the performance ranking of embedding models on both the general MTEB and FinMTEB datasets, calculating Spearman’s rank correlation between the two. The results, shown in Table  \ref{tab: ranking}, indicate that the ranking correlation is not statistically significant (p-values all greater than 0.05). In other words, a general-purpose embedding model performing well on MTEB does not necessarily perform well on domain-specific tasks.
\label{append: spearman}
\begin{center}

    \begin{table*}[ht]
    \centering
    \resizebox{0.7\linewidth}{!}{
        \begin{tabular}{lccccccc}
        \toprule
        & \textbf{STS} & \textbf{Class.} & \textbf{Ret.} & \textbf{Rerank.} & \textbf{Clust.} & \textbf{PairClass.} & \textbf{Summ.} \\
        \midrule
        \textbf{Correlation} & 0.30 & -0.80 & 0.30 & -0.10 & -0.70 & -0.30 & 0.60 \\
        \textbf{p-value} & 0.62 & 0.10 & 0.62 & 0.87 & 0.18 & 0.62 & 0.28 \\
        \bottomrule
        \end{tabular}
    }
    \caption{Spearman’s correlation of embedding models’ performance on MTEB and FinMTEB across different tasks. The p-value indicates that all correlations are statistically insignificant, suggesting a lack of evidence for a relationship between embedding model performance on the two benchmarks.}
    \label{tab: ranking}
    \end{table*}

\end{center}

\section{Analysis of Variance (ANOVA)}

\label{append:anova}
Table \ref{tab:anova_results} illustrates the full results of ANOVA analysis.
\begin{table*}[htbp]
\centering
\resizebox{.9\textwidth}{!}{
\begin{tabular}{lccccc}
\toprule
\textbf{Task} & \textbf{Factor} & \textbf{Sum of Squares} & \textbf{Degrees of Freedom} & \textbf{F-Statistic} & \textbf{p-value} \\
\midrule
\multirow{3}{*}{\textbf{Classification}} 
 & Model Factor    & 4.17     & 6.00    & 25.55      & $3.41 \times 10^{-30}$ \\
 & Domain Factor   & 56.82    & 1.00    & 2086.30    & $\approx 0$ \\
 & Residual        & 190.42   & 6992.00 & NA         & NA \\
\midrule
\multirow{3}{*}{\textbf{Retrieval}} 
 & Model Factor    & 104.25   & 6.00    & 9052.57    & $\approx 0$ \\
 & Domain Factor   & 6.16     & 1.00    & 3207.72    & $\approx 0$ \\
 & Residual        & 13.42    & 6992.00 & NA         & NA \\
\midrule
\multirow{3}{*}{\textbf{STS}} 
 & Model Factor    & 10.55    & 6.00    & 149.00     & $1.64 \times 10^{-178}$ \\
 & Domain Factor   & 304.09   & 1.00    & 25761.71   & $\approx 0$ \\
 & Residual        & 82.53    & 6992.00 & NA         & NA \\
\midrule
\multirow{3}{*}{\textbf{Clustering}} 
 & Model Factor    & 0.29     & 6.00    & 47.60      & $1.59 \times 10^{-57}$ \\
 & Domain Factor   & 32.25    & 1.00    & 32161.37   & $\approx 0$ \\
 & Residual        & 7.01     & 6992.00 & NA         & NA \\
\midrule
\multirow{3}{*}{\textbf{Summarization}} 
 & Model Factor    & 12.98    & 6.00    & 145.31     & $2.90 \times 10^{-174}$ \\
 & Domain Factor   & 14.49    & 1.00    & 973.32     & $3.60 \times 10^{-200}$ \\
 & Residual        & 104.07   & 6992.00 & NA         & NA \\
\midrule
\multirow{3}{*}{\textbf{Reranking}} 
 & Model Factor    & 5.38     & 6.00    & 489.05     & $\approx 0$ \\
 & Domain Factor   & 0.64     & 1.00    & 346.78     & $1.39 \times 10^{-75}$ \\
 & Residual        & 12.84    & 7002.00 & NA         & NA \\
\midrule
\multirow{3}{*}{\textbf{Pair Classification}} 
 & Model Factor    & 0.25     & 6.00    & 1.97       & 0.07 \\
 & Domain Factor   & 249.19   & 1.00    & 11989.92   & $\approx 0$ \\
 & Residual        & 145.31   & 6992.00 & NA         & NA \\
\midrule
\textbf{Average} 
 & Model Factor    & 0.00     & 6.00    & 1.34       & 0.37 \\
 & Domain Factor   & 0.08     & 1.00    & 253.87     & $\approx 0$ \\
 & Residual        & 0.00     & 6.00    & NA         & NA \\
\bottomrule
\end{tabular}
}
\caption{
\textbf{Analysis of Variance (ANOVA) Results Across Tasks and Factors.} 
\textit{Factor} represents the independent variables analyzed: \textbf{Model Factor} pertains to variations attributed to different models, and \textbf{Domain Factor} pertains to variations due to different domains (MTEB or FinMTEB). \textbf{Residual} refers to the unexplained variance. The \textbf{Sum of Squares}, \textbf{Degrees of Freedom}, \textbf{F-Statistic}, and \textbf{p-value} are presented for each factor within each task. Asterisks denote significance levels, with lower p-values indicating higher statistical significance. The Domain Factor consistently shows high significance across all tasks.
}
\label{tab:anova_results}
\end{table*}

\section{Generic vs. Finance-specific Model Comparison}
\label{app:generic_vs_finance}

Table~\ref{tab:fin_e5_vs_baseline_significance} presents paired t-test results comparing finance-adapted models with their generic baselines. Both comparisons show clear performance gains from domain-specific training. Fin-E5 achieves statistically significant gains in Retrieval (p = 0.0489) and Classification (p = 0.0206) tasks. FinBERT shows significant improvements in Retrieval (p = 0.0472), Classification (p = 0.0009), and Reranking (p = 0.0093) tasks. These improvements are particularly valuable since retrieval and classification underpin core financial workflows such as document search and risk assessment.

\begin{table*}[htbp]
\centering
\begin{tabular}{lcccccccc}
\toprule
\multirow{2}{*}{\textbf{Task}} & \multirow{2}{*}{\textbf{Datasets}} & \multicolumn{3}{c}{\textbf{Fin-E5 vs. E5-mistral}} & \multicolumn{3}{c}{\textbf{FinBERT vs. BERT}} \\
\cmidrule(lr){3-5} \cmidrule(lr){6-8}
& & \textbf{Fin-E5} & \textbf{E5-mistral} & \textbf{p-value} & \textbf{FinBERT} & \textbf{BERT} & \textbf{p-value} \\
\midrule
STS & 2 & 0.4342 & 0.3800 & 0.1252 & 0.4198 & 0.3790 & 0.4681 \\
Retrieval & 10 & 0.7105 & 0.6749 & 0.0489* & 0.1102 & 0.0207 & 0.0472* \\
Classification & 8 & 0.7565 & 0.6449 & 0.0206* & 0.5923 & 0.5496 & 0.0009* \\
Clustering & 6 & 0.5650 & 0.5783 & 0.1864 & 0.2833 & 0.1744 & 0.0732 \\
Reranking & 3 & 0.9896 & 0.9875 & 0.1623 & 0.6404 & 0.3930 & 0.0093* \\
PairClassification & 3 & 0.8014 & 0.7394 & 0.2066 & 0.6967 & 0.7111 & 0.2431 \\
Summarization & 3 & 0.4797 & 0.5275 & 0.3607 & 0.2010 & 0.1686 & 0.2138 \\
\bottomrule
\end{tabular}
\caption{Performance comparison between finance-adapted models and their generic baselines. * indicates statistical significance at $\alpha$ = 0.05.}
\label{tab:fin_e5_vs_baseline_significance}
\end{table*}


\end{document}